\documentclass[conference]{IEEEtran}
\IEEEoverridecommandlockouts

\usepackage{cite}
\usepackage{amsmath,amssymb,amsfonts}
\usepackage{algorithmic}
\usepackage{graphicx}
\usepackage{textcomp}
\usepackage{xcolor}
\usepackage{url}
\usepackage{booktabs}
\usepackage{array}
\usepackage{tabularx}
\newcolumntype{Y}{>{\raggedright\arraybackslash}X}
\def\BibTeX{{\rm B\kern-.05em{\sc i\kern-.025em b}\kern-.08em
    T\kern-.1667em\lower.7ex\hbox{E}\kern-.125emX}}

\usepackage[most]{tcolorbox}

\begin{document}

\title{What Social Media Teaches Us About Accumulative AI Risks}

\author{
\IEEEauthorblockN{Dawei Li, Bohan Jiang, Huan Liu}
\IEEEauthorblockA{
School of Computing and Augmented Intelligence\\
Arizona State University\\
\{daweili5, bjiang14, huanliu\}@asu.edu
}
}

\maketitle

\begin{abstract}
Prominent AI safety discussions emphasize existential threats and catastrophic events, such as AI-enabled nuclear conflict or engineered biological outbreaks. This position paper argues that there is another type of AI risk that arises and accumulates from the widespread use of AI. Drawing on lessons from social media, we argue that safety assessments must also address accumulative harms that develop through repeated use, recurring decisions, and feedback across interconnected deployments. Repeated reassurance, selective exposure, recurring decision biases, and growing personal profiles may gradually narrow perspectives, weaken independent judgment, or enable exploitative influence without a single clearly harmful event. We advance the Trajectory Neglect Thesis: a focus on vivid catastrophic events and identifiable incidents risks overlooking less visible harms that accumulate through the widespread daily use of AI. Social media research motivates this position by showing how reciprocal media effects, social reinforcement, and individual susceptibility shape outcomes beyond individual encounters. We examine how related mechanisms may operate in AI systems, using conversational AI as a concrete setting where generation, semantic inference, and persistent memory may tighten feedback loops. Sycophancy, manipulation, and privacy loss provide a case study of how these risks may interact and accumulate. This comparison identifies mechanisms to investigate rather than assuming that repeated AI use is harmful. We therefore call for evaluation of cumulative patterns of use and deployment, individual and societal outcomes, careful causal inference, and proportionate safeguards that preserve useful assistance and informed user choice.
\end{abstract}


\section{Introduction}
\label{sec:introduction}


Prominent AI safety discussions emphasize existential threats and catastrophic events~\cite{gyevnar2025ai}. For example, an advanced AI system could contribute to nuclear escalation, enable the release of an engineered pathogen with pandemic potential, or trigger cascading failures across critical infrastructure. The possibility of widespread loss of life and irreversible societal damage makes these risks particularly salient in public debate. Scenarios such as nuclear conflict and pandemics also provide concrete, familiar ways to communicate the danger. Their potential severity and visibility help explain why existential threats attract substantial public attention. This position paper argues that there is another type of AI risk that arises and accumulates from the widespread daily use of AI. These accumulative harms are harder to assess because they may develop through repeated use, recurring decisions, or feedback across interconnected deployments without any single event clearly violating such a requirement. These processes may gradually change users' beliefs and reliance, reinforce biases in institutional decisions, or expand the collection and reuse of personal information. Preventing identifiable failures, therefore, leaves a further safety question: how can individually acceptable AI outputs and actions contribute to severe, harmful outcomes over time?

Lessons from social media bring this question into focus. Media-effects research has long treated people as active selectors whose experiences can shape later choices~\cite{slater2007reinforcing}. Work on social learning shows how feedback can alter future online expression~\cite{brady2021social}. Other studies find substantial differences in the associations between media use and well-being across individuals~\cite{beyens2020effect}. Together, these findings direct attention to a process in which exposure affects a user's response, and that response helps shape what the user encounters next. The consequences of this process depend on the sequence of interactions and the user's circumstances, which cannot be fully captured by judging each piece of content separately.

Consider someone who repeatedly asks an assistant whether a disagreement proves that others cannot be trusted. Each response may acknowledge distress, offer a plausible interpretation, and avoid prohibited content. Across sessions, however, the assistant may keep validating the same account, learn which explanations the user accepts, and offer increasingly tailored reassurance. The relevant concern is the direction of the exchange: whether it encourages reflection or gradually makes alternatives harder to consider. This is an illustrative risk scenario, not evidence that such a trajectory occurs in every system. It shows why an output can be locally acceptable while the pattern deserves investigation.

Building on broader work on accumulative risk~\cite{kasirzadeh2025two,Li2025BeyondSA}, we use \emph{accumulative AI risk} to describe risk that develops through the repetition, ordering, persistence, or feedback of user--AI interactions. Their importance depends on what changes over time, for whom, and under which conditions. Repetition can also strengthen learning, provide useful support, and help users pursue their own goals. The safety question is therefore not whether an interaction is long. It is whether its trajectory creates or increases the likelihood of a specified adverse outcome.

Conversational AI changes the setting in which these mechanisms operate. A system can respond to what a user says, infer meaning from the conversation, and adapt its next answer. Where memory is enabled~\cite{tan2025prospect}, information from one session may shape another. A general-purpose assistant can also connect advice across topics that users previously encountered through separate services or relationships. These features create plausible routes for accumulative influence. Social media provides a guide to investigating these routes, rather than proof that the two technologies produce identical effects.

Based on these observations, we propose that accumulative AI risks deserve sustained attention, just as existential AI risks do. Social media research shows how repeated exposure and feedback can shape beliefs, behavior, and later choices in ways that individual encounters do not reveal. The widespread daily use of AI creates opportunities for similar processes to affect users and institutions over time. Even modest effects may become consequential when they persist, reinforce one another, or extend across large populations. Their gradual development makes them harder to recognize and attribute, but lower visibility does not establish lower significance. These mechanisms provide a reason to take accumulative risks seriously before they produce an unmistakable crisis.

The paper develops its argument through four successive steps:
\begin{itemize}
    \item First, we identify an asymmetry in risk visibility. Existential threats framed around a specific catastrophic event are easier to imagine and communicate, while accumulative harms arising from widespread, repeated AI use may be less visible and receive less attention.

    \item Second, we draw on social media research to explain why this gap matters. Repeated exposure, feedback loops, and differences across users show why acceptable individual content does not guarantee safe user trajectories.

    \item Then, we examine how these lessons apply to conversational AI. We map relevant mechanisms, compare differences between the two settings, and use sycophancy, manipulation, and privacy loss as a case study of how risks may interact and accumulate.

    \item Finally, we derive implications for evaluation and deployment. Assessments should track changes across interactions, while safeguards should target supported mechanisms and preserve useful assistance and informed user choice.
\end{itemize}

\section{The Asymmetry of Risk Visibility}
\label{sec:asymmetry}

\subsection{Catastrophic Events and Accumulative Trajectories}

Existential risk concerns consequences severe enough to threaten humanity's survival or future potential. When framed around a decisive event, such as nuclear conflict or a catastrophic biological outbreak, it provides a concrete scenario through which people can imagine the danger. The event and its consequences offer a clear focus for discussion, even when their likelihood remains difficult to estimate. Accumulative risks may lack an equally recognizable focal point: harm can develop through many uses and decisions without a single event that makes the overall process apparent.

An identifiable incident, such as an unauthorized disclosure or the provision of a dangerous instruction, similarly supplies a relatively clear starting point for investigation. Its consequences may be delayed, but the event can be examined. A trajectory-based accumulative risk instead depends on a pattern over time: what is repeated, what is omitted, how an earlier action changes a later decision, and whether previously collected information remains influential. These are different ways to examine risk, not mutually exclusive categories. A trajectory may contain recognizable incidents, and accumulated effects may contribute to a later catastrophe.

An \emph{AI-use trajectory} includes the sequence of uses or decisions, their timing and context, relevant system adaptations, and changes in the people or institutions involved. In conversational AI, an interaction trajectory includes exchanges, their timing and context, relevant system adaptations, and changes in the user's experience or behavior. It may span several sessions. The interaction history records part of this trajectory, but it does not show everything that matters. The user's interpretation, other sources of advice, and events outside the system can affect what follows. Longer histories are therefore useful evidence, not a substitute for studying users and their circumstances.

Accumulation is also more than a count of small failures. Repeating an answer can change its apparent credibility. A request for information can become more persuasive after trust develops. Details shared for separate tasks can become more revealing when combined. In institutional settings, repeatedly relying on similar AI recommendations may reinforce a decision bias across many cases. The same actions may have different consequences in a different order. A later response or intervention can also correct an earlier mistake or weaken a harmful pattern.

\subsection{Why Accumulative Risks Are Harder to See}

The asymmetry has three sources. Salience and observability are the first. A catastrophic scenario provides a recognizable event and severe consequences that are easy to communicate. Accumulative harms may emerge without a comparable moment that attracts attention. At the level of deployed systems, a dangerous answer can be copied and inspected. A gradual shift in a user's confidence, willingness to consider other views, or reliance on an assistant may leave no equally clear artifact. An apparently supportive reply can look appropriate to an outside reviewer while contributing to a pattern the reviewer cannot see.

Attribution is the second. Users bring prior beliefs, needs, and relationships into an interaction. These factors influence both what they ask and how they respond. If a user becomes more isolated while using an assistant, the assistant may be a contributor, a response to isolation, or both. Similar difficulties arise when institutional outcomes reflect many AI-assisted decisions alongside organizational practices and other influences. Identifying reciprocal influence requires suitable comparisons and repeated observations. A temporal association alone does not establish causation.

Measurement is the third. Output tests often have a clear target behavior and repeatable scoring procedure. Changes in user experience or institutional practice require choices about outcome, duration, baseline, and meaningful variation. Different users and settings may follow different paths. An average can conceal a harmed subgroup, but subgroup analysis can also produce unreliable findings unless it is well designed. These difficulties increase the effort needed to detect accumulative risk; they are not evidence that the risk is absent.

The asymmetry matters for attention and decisions. A vivid catastrophic scenario is easier to communicate, while a failure with a clear record is easier to reproduce, compare, and assign to a team. A concern requiring extended observation may struggle to attract comparable attention or enter the same workflow. Where an organization treats what is easily noticed or counted as a complete account of safety, it creates a blind spot. We argue that uncertainty should prompt an appropriate study design, rather than automatically lowering the concern's priority.

\subsection{The Trajectory Neglect Thesis}

\begin{tcolorbox}[
    colback=blue!5!white,
    colframe=blue!30!white,
    boxrule=0.5pt,
    arc=2pt,
    left=6pt,
    right=6pt,
    top=6pt,
    bottom=6pt
]
\textbf{Definition (Trajectory Neglect).}
\emph{Trajectory neglect refers to the underestimation of harms
that depend on histories of AI use or deployment when safety
discussions and assessments prioritize vivid catastrophic events,
isolated outputs, or identifiable incidents.
These harms may be less visible without being less significant.}
\end{tcolorbox}

The thesis does not rank all accumulative risks above existential threats or assume that they have the same severity. It argues that their lower visibility should not justify less serious attention. It also specifies an inference that safety work should avoid: concluding that a system is safe for sustained use because no unacceptable response has been found in the sampled outputs. Such a conclusion needs evidence about sustained use. This remains true when the sample contains several conversation turns but the assessment only labels the final answer.

Nor is every change in a user evidence of harm. Advice is often intended to change beliefs or actions. A useful assistant may help someone become more confident, manage a difficult task, or feel less lonely. Studies of AI companions, for example, report momentary reductions in loneliness following use across a week~\cite{de2026ai}. To identify harm, an assessment must specify an adverse outcome and examine its relationship to AI use. The normative standard and the empirical evidence should both be open to scrutiny.

This distinction turns visibility into a question of evidence coverage. Output checks establish certain properties of responses. Trajectory studies establish other properties of use and deployment. Neither can simply stand in for the other. The practical implication is to ask which consequences a safety assessment can detect and which remain outside its design. Accumulative risks deserve sustained investigation alongside existential threats, even when they do not present an equally vivid catastrophic scenario.

\section{Lessons from Social Media: When Content Safety Misses Trajectory Harm}
\label{sec:socialmedia}

Social media offers a valuable reference because it shows how repeated, personalized interactions can generate risks that are difficult to detect in individual encounters. Its experience helps identify mechanisms to examine in conversational AI, including reinforcement, behavioral adaptation, and feedback loops. The value of this comparison lies in guiding what to measure over time, without assuming that the two technologies produce identical effects.

\subsection{Safety Depends on What Users See, How Often, and in What Order}

Social media safety often focuses on detecting threats, abuse, and misinformation. These checks are necessary, but they do not capture the full user experience. A feed determines which posts users see, how often they see them, and in what order. Even acceptable posts can create concerns when a feed repeatedly encourages social comparison or presents only a narrow range of views.

Research on echo chambers shows that users with similar views often cluster together and that information spreads selectively across these networks~\cite{cinelli2021echo}. This does not mean that every such network radicalizes its members. It shows why checking individual posts is insufficient: every post may be permissible while a user's exposure remains persistently narrow. Detecting that pattern requires examining repeated exposure and the network that shapes it.

Social feedback can also influence future behavior. ~\cite{brady2021social} found that positive feedback on expressions of moral outrage increased the likelihood of later outrage expression. Users also adapted their expression to the norms of their networks. These findings concern what users express, rather than directly measuring how angry they feel. The safety lesson is that feedback can shape the next round of content, so evaluation should examine both the content and the process that encourages its production.

\subsection{User Choices and Media Effects Shape Each Other}

The reinforcing-spirals perspective explains how content selection and media effects can influence each other over time~\cite{slater2007reinforcing}. People choose content based partly on their existing interests and views. What they encounter may then change those interests and views, influencing what they choose next. Other experiences and social conditions can strengthen or weaken this process; it does not always escalate.

This means that repeated selection is not necessarily evidence of a fixed, independent preference. It may reflect an existing preference, the influence of earlier exposure, or both. The same question applies to AI: a user's next prompt may partly reflect the assistant's previous replies. Evaluation should therefore consider how the interaction shapes later requests, rather than treating every request as independent of what came before.

Responsibility is also shared. Users make choices, but platforms influence the options they encounter and the feedback they receive. Ranking systems and engagement objectives can shape these patterns without anyone intending a harmful outcome. For AI assistants, response style, memory, and deployment objectives may similarly influence repeated use. An adverse trajectory can emerge even when both the user and the system are pursuing apparently reasonable goals.

\subsection{Average Effects Can Hide Different User Outcomes}

The Differential Susceptibility to Media Effects Model explains why media effects vary with personal characteristics, developmental stage, and social circumstances~\cite{valkenburg2013differential}. The same content or amount of use can affect people differently. Time spent online alone therefore provides an incomplete measure of risk.

\cite{beyens2020effect} found that the relationship between social media use and momentary well-being varied substantially across adolescents. The association was positive for some, negative for others, and small or absent for others. These findings do not establish a general causal effect, but they show why population averages should be accompanied by analysis of changes within individual users.

AI evaluation should ask similar questions. For example, an assistant might help one user reflect on a concern while encouraging another to repeatedly seek reassurance. This is a hypothesis to test, not a diagnosis to infer from conversation alone. The goal is to identify when particular system behaviors, user circumstances, and patterns of use lead to beneficial or adverse outcomes.




The practical lesson is to test outcomes directly. Acceptable individual posts do not guarantee safe patterns of repeated use. At the same time, repetition and personalization alone do not establish harm. Safeguards should address mechanisms supported by evidence and be evaluated in the settings where they will be used.

Social media is therefore a valuable reference for conversational AI. It provides evidence and research methods for studying repeated exposure, behavioral reinforcement, and differences across users. These lessons help identify what to measure over time, without assuming that conversational AI will produce the same effects.

\section{From Social Media to Conversational AI}
\label{sec:transfer}

Research on social media offers useful lessons for studying risks in conversational AI, but it does not establish that repeated AI interaction is necessarily harmful. In this section, we map relevant mechanisms from social media to conversational AI, compare the differences that may change how these mechanisms operate, and use a case study to illustrate how accumulative risks may arise through repeated interaction.

\subsection{Mechanism Transfer: Observation, Inference, and Influence}

The comparison rests on a shared process. A service observes behavior, infers something about the user, adapts what it presents, and receives another response. That response can affect both the system's next action and the user's next choice. Table~\ref{tab:mechanisms} maps these steps across feeds and conversations. It is a conceptual comparison, not evidence that the resulting harms have the same magnitude.

\begin{table*}[t]
\caption{A conceptual mapping from social-media feeds to conversational AI. The final column identifies what a trajectory assessment should examine; it does not assert a proven adverse effect.}
\label{tab:mechanisms}
\centering
\footnotesize
\begin{tabularx}{\textwidth}{@{}p{0.12\textwidth}YYY@{}}
\toprule
\textbf{Mechanism} & \textbf{Social-media example} & \textbf{Conversational-AI counterpart} & \textbf{Trajectory question} \\
\midrule
Observation & Clicks, follows, dwell time, and reactions & Requests, corrections, disclosures, and return visits & Which signals shape later interactions? \\
Inference & Estimates of interests or likely engagement & Interpretation of goals, beliefs, and personal circumstances & Which inferred assumptions persist or change? \\
Personalization & Ranking and selection of available content & Tailored framing, explanations, questions, and advice & Does adaptation broaden options or narrow them? \\
Influence & Exposure, social feedback, and visible network norms & Reassurance, recommendations, and responsive dialogue & What changes in judgment, behavior, or reliance? \\
Updating & Revised ranking signals and renewed user selection & Context or memory updates and the user's next request & Does feedback correct, maintain, or amplify a pattern? \\
\bottomrule
\end{tabularx}
\end{table*}

Observation differs in form. A feed can learn from choices among visible items. An assistant receives accounts of what the user is worried about, wants to do, or believes has happened. It can also ask a question that elicits additional information. This makes disclosure part of the feedback process, not merely data that precedes it.

Inference can extend beyond what a person intends to reveal. Research has shown that digital records can predict sensitive personal attributes~\cite{kosinski2013private}, and that language models can infer private attributes from ordinary text~\cite{staab2024beyond}. We infer that a conversation may provide a richer basis for later adaptation as details are combined. An observed inference capability does not mean that every assistant builds such a profile, but it is relevant wherever a deployment permits the information to be used.

Influence also changes form. A feed primarily distributes and orders content, although social media also supports direct interaction. A conversational system can produce an explanation addressed to the user's immediate objection. The user's acceptance, resistance, or clarification can then shape the next answer. The mechanism is not restricted to political persuasion. It can operate in repeated advice seeking, interpretation of social events, and decisions about what information to share.

\subsection{Feedback-Loop Tightening}

We use feedback-loop tightening to describe a closer connection between observation, adaptation, and renewed influence. This is a hypothesis about design affordances, not a demonstrated comparison of causal effect sizes. Four features make it plausible.

\textbf{Generation} allows the system to alter the response itself, rather than only choose among existing items. It can change emphasis, supply a new justification, or restate an argument in language that fits the conversation. Ranking and generation can coexist in both technologies; the distinctive concern is how much of the next encounter can be shaped around the user's latest response.

\textbf{Semantic inference} allows explicit statements to become cues for adaptation. A user may explain why a suggestion was unwelcome, describe a relationship, or identify a fear. Such signals can help an assistant give better advice. They can also make it easier to repeat a framing the user already accepts or identify a point of vulnerability. The risk depends on the system's objective and how the information is used.

\textbf{Persistent memory}, where enabled, can extend the process across sessions. A useful preference remembered yesterday may shape today's recommendation. A mistaken assumption can persist in the same way. The important distinction is between adapting within the current context, retrieving information stored about the user, and changing model parameters. These are different processes. Our argument does not assume that every assistant learns new model weights from each conversation.

\textbf{Universality}, understood here as use across many domains, can connect otherwise separate contexts. The same assistant may help with work, relationships, shopping, and emotional support. If information is carried across these roles, an inference formed in one setting can shape influence in another. This capability can improve continuity. It can also make the boundaries of information use and advice harder for users to track.

Tightening may reduce outside visibility as well. A public post can be inspected by many people, whereas a tailored exchange may be visible mainly to its participants and provider. Reviewers who sample replies without the relevant history can miss how earlier assumptions are reused. The user sees the conversation, but may not see the stored profile or objective guiding adaptation. These conditions motivate greater attention to context.

\subsection{A Case Study in Sycophancy, Manipulation, and Privacy Loss}

We use SMP for sycophancy, manipulation, and privacy loss. These describe three distinct concerns: undue agreement, covert or exploitative influence, and diminished control over personal information. Their interaction illustrates how risks may accumulate across a conversation, even when individual replies appear acceptable.

Sycophancy can repeatedly reinforce a user's beliefs. Research finds that preference-based training can favor answers that match users' beliefs over truthful alternatives~\cite{sharma2024towards}. The concern extends beyond whether one answer is false. Does the assistant repeatedly accept the user's interpretation without sufficient justification? Does this pattern change how the user seeks or considers other perspectives?

A recent work by ~\cite{ibrahim2026relational} reports that repeated interaction with sycophantic AI over three weeks affected advice seeking and satisfaction with real-world social interactions. These findings apply to the response styles and population studied. They nevertheless illustrate why feeling understood in one exchange may not capture the consequences of repeated interaction.

Manipulation concerns influence that exploits vulnerabilities or conceals its purpose, undermining autonomous choice~\cite{susser2019technology}. Persuasion alone is not manipulation: users can benefit from changing their views in response to good reasons. The relevant questions are whether an assistant repeatedly hides alternatives, exploits concerns disclosed earlier, or steers the user toward an undisclosed objective.

Privacy loss concerns diminished control over personal information as it is collected, inferred, retained, or reused beyond what the user understands or authorizes. Repeated conversation can bring together details that were disclosed separately, producing a more revealing account of the user. Disclosure itself does not establish privacy loss. The concern is whether the resulting information is used in ways the user neither anticipates nor meaningfully controls.

Personal information may also provide a means for targeted influence. A controlled debate study found that personalized GPT-4 was highly persuasive relative to human opponents~\cite{salvi2025conversational}. The study does not establish privacy loss or a separate causal effect of personalization, but it motivates examining how conversational systems use personal information when influencing users.

Consider a user discussing repeated interpersonal conflict. Agreeable replies encourage the user to return and share increasingly sensitive details. Across conversations, these disclosures reveal relationships, insecurities, and recurring concerns. If the system retains or reuses this information beyond the user's understanding or authorization, privacy loss can occur. If it then exploits those concerns to narrow the user's choices toward a concealed objective, manipulation can follow. Sycophancy may help sustain the interaction that enables both. This is a possible mechanism to test; the cited studies do not establish a complete SMP feedback loop.

A conventional content check may approve many replies while missing repeated endorsement, cumulative disclosure, and later exploitation of personal information. Evaluating SMP therefore requires following how the assistant's responses, the user's behavior, and the use of disclosed information change over time. A useful comparison would examine assistance that remains supportive while preserving justified uncertainty, informed choice, and user control over personal information.

\section{Toward Trajectory-Centered AI Safety}
\label{sec:agenda}

\subsection{Evaluate Histories and the Conditions That Produce Them}

Evaluation should match the mechanism being claimed. A privacy incident may require inspecting one disclosure. A concern about persistent reassurance or tailored steering requires histories in which a user's requests and the system's adaptations affect later exchanges. Relevant test variations include repetition, the order of disclosures, disagreement, pauses between sessions, memory settings, and changes of topic. Testing should establish whether the system maintains a risky pattern, corrects it, or changes course when the user expresses a different goal.

History-based evaluation can use controlled scenarios to compare response policies. For example, evaluators can ask whether a system gradually becomes more certain about an unresolved interpretation, reuses an inferred vulnerability outside its original context, or repeatedly suppresses relevant alternatives. These are behavioral questions with inspectable evidence. Synthetic users can help explore such patterns, but simulated distress, trust, or dependence does not establish those outcomes in people. Claims about human consequences need human evidence.

Evaluation should also include the deployed configuration. Memory defaults, system instructions, information access, and objectives can change what a fixed model does across sessions. Providers should report which configuration was evaluated and which contexts remain untested. A good output benchmark result should support claims within its scope, rather than become an assurance about every form of sustained use.

\subsection{Track User Outcomes and Separate Association from Causation}

Longitudinal studies should distinguish system behavior, interaction patterns, and user outcomes. Repeated endorsement is a behavioral signal. Increased advice seeking is a change in use. Reduced satisfaction with human relationships is a reported outcome. These measures can inform one another, but they are not interchangeable. A responsible study should specify what would count as an adverse change and why, alongside potential benefits.

Existing research illustrates the need for this separation. Fang and colleagues studied extended chatbot use over four weeks, comparing interaction modes and conversation types~\cite{fang2025psychosocial}. The updated report found no significant effects of the assigned conditions on the measured outcomes, while more voluntary use and some user characteristics were associated with worse outcomes. Because usage intensity was not itself randomized, those associations do not establish that additional use caused the changes. This distinction is central to trajectory research, not a reason to discard it.

Studies should collect suitable baselines, repeat assessments, examine variation across users, and report the limits of their follow-up period. Where feasible, comparisons can vary a specific response policy or memory practice while keeping other features similar. Follow-up beyond a session can help distinguish momentary relief from more durable outcomes. Findings should be described at the level the design supports: an observed pattern, a supported causal effect in a setting, or a hypothesis requiring further study.

An assessment of agency also needs more than measuring whether a user accepted advice. It can examine whether users understood a recommendation's purpose, considered meaningful alternatives, and could decline or revise it. Well-being measures should be chosen with appropriate expertise and participant input. No single score can resolve all questions about a user's interests, and automated labels should not be treated as diagnoses.

\subsection{Detect Patterns Early and Use Proportionate Safeguards}

Early detection should focus on supported patterns rather than declaring all sustained engagement risky. Candidate signals include repeated unwarranted endorsement, increasingly narrow recommendations, recurring use of an inferred vulnerability, or personal information carried into an unrelated role. These are proposed indicators for validation. A detector's false alarms and missed cases matter, as do the effects of the intervention it triggers.

The response should fit the pattern. Repeated confirmation may call for preserving uncertainty and inviting relevant alternatives. Persistent use of a mistaken assumption may call for asking the user to confirm or correct it. Cross-context reuse of personal information may call for a clear permission boundary, visible memory controls, or restricting that reuse. A concealed incentive behind advice requires disclosure or a change in the system's objective. Restricting conversation length alone would not address all of these mechanisms.

Interventions also require evaluation. The social media evidence cautions against assuming that exposure to a competing view necessarily improves outcomes~\cite{bail2018exposure}. An assistant should not automatically become confrontational whenever a conversation is repetitive. Useful support can remain available while the system makes the limits of its knowledge and the basis of its advice understandable. More restrictive measures need stronger evidence of likely harm and attention to the costs for users who benefit from the service.

Trajectory safety must not become a justification for unlimited surveillance. Detecting a pattern can itself require sensitive history. Data minimization, consent, limited retention, and secure access should be part of the evaluation design. Where possible, user-visible controls and narrowly scoped processing should provide protection without constructing a comprehensive psychological profile. A safety mechanism that expands unrelated information collection can create a new risk while attempting to address another.

\subsection{Make Longitudinal Evidence Part of Deployment Decisions}

Providers should state what their safety evidence covers: tested outputs, tested interaction patterns, study populations, configurations, and follow-up periods. For products intended for recurring personal advice or companionship, the evidence should include an account of sustained use proportionate to the role being offered. This obligation does not require proof that every possible long-term harm has been eliminated. It requires recognizing which claims an output test cannot support.

Independent study is also important. Researchers need appropriately governed access to evidence about repeated use, while users need clear ways to review relevant memory, challenge an inference, and report a harmful pattern. Reports should be able to describe a sequence without forcing the user to identify one offending sentence. The National Academies' recommendations for longitudinal research and appropriate platform data access provide a relevant precedent~\cite{nationalacademies2024social}.

The main objection is that trajectory research is costly and may delay beneficial systems. The answer is to scale the evidence to the deployment claim and plausible mechanism. A bounded tool for an isolated task need not receive the same evaluation as a persistent personal adviser. Reversible safeguards can be tested while stronger evidence develops. We propose a broader evidentiary standard for sustained interaction, with transparent limits, rather than a requirement for certainty before any release.

\section{Conclusion}
\label{sec:conclusion}

Our position is that AI safety must also address accumulative risks that arise from the widespread daily use of AI. Social media teaches that consequences cannot be inferred from individual pieces of content alone. Exposure, feedback, user selection, and susceptibility shape outcomes over time. Related processes may operate through repeated AI use, recurring decisions, and interconnected deployments. Conversational AI illustrates how tailored replies, inferred personal information, and persistent memory can sustain these processes.

The Trajectory Neglect Thesis highlights why such risks deserve attention alongside existential threats: harms that develop gradually may remain overlooked without a vivid catastrophe or clearly harmful output. Their lower visibility does not establish lower significance. Acceptable individual outputs are therefore insufficient evidence for safe sustained use. Evaluation, research, and safeguards should examine how effects accumulate, identify adverse outcomes for users and institutions, and preserve the benefits of AI.

\section*{Acknowledgement}

This work was supported by the U.S. Army Materiel Command under Grant Award Number W911NF-24-2-0175 and by the U.S. Army Research Laboratory under Grant Award Number W911NF2020124.

\bibliographystyle{abbrv}
\bibliography{IEEEabrv}

\end{document}